\pdfoutput=1
\documentclass[11pt]{article}
\usepackage{syntaxfest2021}
\usepackage{times}
\usepackage{url}
\usepackage{latexsym}

\usepackage{amsmath}
\usepackage{amssymb}
\usepackage{graphicx}
\usepackage{transparent}
\usepackage{subcaption}
\usepackage{booktabs}
\usepackage[utf8]{inputenc}
\usepackage[T1]{fontenc}
\usepackage{multirow}
\usepackage{pdfcomment}
\usepackage{fontawesome5}

\newcommand{\class}{\textsc{Class}}
\newcommand{\boot}{\textsc{Boot}}
\newcommand{\zero}{\textsc{Zero}}
\newcommand{\lda}{\textsc{LDA}}
\newcommand{\gmm}{\textsc{GMM}}
\newcommand{\ldal}{\textsc{LDA+l}}
\newcommand{\gmml}{\textsc{GMM+l}}
\newcommand{\freq}{\textsc{Freq}}

\makeatletter
\newif\ifnobrackets
\renewcommand\@cite[2]{\ifnobrackets\else(\fi{#1\if@tempswa , #2\fi}\ifnobrackets\else)\fi\nobracketsfalse}
\newcommand\citealp{\nobracketstrue\cite}
\makeatother

\syntaxfestfinalcopy 

\title{How Universal is Genre in Universal Dependencies?}

\author{Max M{\"u}ller-Eberstein \and Rob van der Goot \and Barbara Plank \\
  Department of Computer Science \\
  IT University of Copenhagen, Denmark \\
  \texttt{mamy@itu.dk, robv@itu.dk, bapl@itu.dk}}

\date{}

\begin{document}
\maketitle

\begin{abstract}
This work provides the first in-depth analysis of genre in Universal Dependencies (UD). In contrast to prior work on genre identification which uses small sets of well-defined labels in mono-/bilingual setups, UD contains 18 genres with varying degrees of specificity spread across 114 languages. As most treebanks are labeled with multiple genres while lacking annotations about which instances belong to which genre, we propose four methods for predicting instance-level genre using weak supervision from treebank metadata. The proposed methods recover instance-level genre better than competitive baselines as measured on a subset of UD with labeled instances and adhere better to the global expected distribution. Our analysis sheds light on prior work using UD genre metadata for treebank selection, finding that metadata alone are a noisy signal and must be disentangled within treebanks before it can be universally applied.
\end{abstract}

\section{Introduction}

%
%
\blfootnote{
    %
    %
    %
    %
    %
    %
    \hspace{-0.65cm}  
    This work is licensed under a Creative Commons 
    Attribution 4.0 International License.
    License details:
    \url{http://creativecommons.org/licenses/by/4.0/}.
}

Identifying document genre automatically has long been of interest to the NLP community due to its immediate applications both in document grouping~\cite{petrenz2012} as well as task-specific data selection~\cite{ruder2017,sato2017}.

Cross-lingual genre identification has however remained a challenge, mainly due to the lack of stable cross-lingual representations~\cite{petrenz2012}. Recent work has shown that pre-trained masked language models (MLMs) capture monolingual genre~\cite{aharoni2020}. Do such distinctions manifest in highly multilingual spaces as well? In this work, we investigate whether this property holds for the genre distribution in the 114 language Universal Dependencies corpus (UD version 2.8; \citealp{ud28}) using the multilingual mBERT MLM~\cite{devlin2019}.

In absence of an exact definition of textual genre~\cite{kessler-etal-1997-automatic,webber-2009-genre,plank2016}, this work will focus on the information specifically denoted by the \texttt{genres} metadata tag in UD. We hope that an in-depth, cross-lingual analysis of what this label represents will enable practitioners to better control for the effects of domain shift in their experiments. Previous work using these UD metadata for proxy training data selection have produced mixed results~\cite{stymne-2020-cross}. We investigate possible reasons and identify inconsistencies in genre annotation. The fact that genre labels are only available at the level of treebanks makes it difficult to gather a clear picture of the \textit{sentence-level} genre distribution --- especially with some treebanks having up to 10 genre labels. We therefore investigate the degree to which instance-level genre is recoverable using only the treebank-level metadata as weak supervision.

Our contributions entail the, to our knowledge, first detailed definition of all UD metadata genre labels (Section \ref{sec:ud-genre}), four weakly supervised methods for extracting instance-level genre across 114 languages (Section \ref{sec:methods}) as well as genre identification experiments which show that our proposed two-step procedure allows for effective genre recovery in multilingual setups where language relatedness typically outweighs genre similarities (Section \ref{sec:experiments}).\footnote{Code available at \href{https://personads.me/x/syntaxfest-2021-code}{\texttt{https://personads.me/x/syntaxfest-2021-code}}.}

\section{Related Work}

The largest hurdle for cross-lingual genre classification is the lack of shared representational spaces. \newcite{sharoff2007} use shared POS n-grams in order to jointly classify the genre of English and Russian documents. \newcite{petrenz2012} similarly seek out features which are stable across languages in order to classify English and Chinese documents into four shared genres. A recent data-driven approach finds that monolingual MLM embeddings can be clustered into five groups closely representing the data sources of the original corpus~\cite{aharoni2020}. In this work, we investigate whether this holds for multilingual settings as well.

Being able to identify textual genre has been crucial for domain-specific fine-tuning~\cite{dai-etal-2020-cost,gururangan2020} including dependency parsing. For parser training, in-genre data is typically selected by proxy of the data source~\cite{plank2011,rehbein-bildhauer-2017-data,sato2017}. Data-driven approaches which include automatically inferred topics based on word and embedding distributions~\cite{ruder2017} as well as POS-based approaches~\cite{sogaard2011,rosa2015,vania2019} have also been found effective.

Universal Dependencies~\cite{nivre-etal-2020-universal} aims to consolidate syntactic annotations for a wide variety of languages and genres under a single scheme. The latest release contains 114 languages --- many with fewer than 100 sentences. In order for languages at all resource levels to benefit from domain adaptation, it will continue to be important to identify cross-lingually stable signals for genre. While language labels are generally agreed upon, differences in genre are more subtle. Metadata at the treebank level provides some insights into genres of original data sources, however these are ``neither mutually exclusive nor based on homogeneous criteria, but [are] currently the best documentation that can be obtained''~\cite{nivre-etal-2020-universal}.

\newcite{stymne-2020-cross} performs an initial study on using these treebank metadata labels for the selection of spoken and Twitter data. Results show that training on out-of-language/in-genre data is superior to out-of-language/out-of-genre data. However the best results are obtained using in-language data regardless of genre-adherence. This holds across multiple methods of proxy dataset selection (e.g.\ treebank embeddings; \citealp{smith2018}).

Recently, \newcite{muellereberstein2021} have shown that combining UD genre metadata and MLM embeddings can improve proxy training data selection for zero-shot parsing of low-resource languages. The use of genre in their work is more implicit as it is mainly driven by the genre of the target data. In contrast, this work takes a holistic view and explicitly examines the classification of instance-level genre for all sentences in UD.

As genre appears to be a valuable signal, we set out to investigate how it is defined and distributed within UD. Due to the coarse, treebank-level nature of current genre annotations, we hypothesize that a clearer picture can only be obtained by moving to the sentence level. We therefore transition from prior supervised document genre prediction to weakly supervised \textit{instance} genre prediction. Additionally, we expand the linguistic scope from mono- or bilingual corpora to all 114 languages currently in UD.

More generally, this task can be viewed as predicting genre labels for all sentences in all corpora of a collection while only being given the set of labels said to be contained in each corpus.

\section{UD-level Genre}\label{sec:ud-genre}

We analyze genre as currently used in the \texttt{genres} metadata of 200 treebanks from Universal Dependencies version 2.8~\cite{ud28}. Section \ref{sec:global-annotations} provides an overview of all UD genre types and Section \ref{sec:instance-annotations} analyzes how these global labels relate to the subset of treebanks which do provide treebank-specific, instance genre annotations.

\subsection{Available Metadata}\label{sec:global-annotations}

UD 2.8~\cite{ud28} contains 18 genres which are denoted in each treebank's accompanying metadata. Around 36\% of treebanks contain a single genre while the remaining majority can contain between 2--10 which are not further labeled at the instance level. There is no official description of each genre label, however they can be roughly categorized as follows:

\nocite{fontawesome}

\paragraph{\indent {\small\faIcon{graduation-cap}} academic} Collections of scientific articles covering multiple disciplines. Note that this label may subsume others such as \textit{medical}.

\paragraph{\indent {\small\faIcon{cloud}} bible} Passages from the bible, frequently from older languages (e.g.\ Old Church Slavonic-PROIEL by \citealp{cu-proiel}). Largely non-overlapping passages are used across treebanks.

\paragraph{\indent {\small\faIcon{calendar}} blog} Internet documents on various topics which may overlap with other genres such as \textit{news}. They are typically more informal in register. Some treebanks group social media content and reviews under this category (e.g.\ Russian-Taiga by \citealp{ru-taiga}).

\paragraph{\indent {\small\faIcon{envelope}} email} Formal, written communication. This includes English-EWT's~\cite{en-ewt} subsection based on the Enronsent Corpus~\cite{styler2011} as well as letters attributed to Dante Alighieri as part of Latin-UDante~\cite{la-udante}.

\paragraph{\indent {\small\faIcon{book}} fiction} Mostly paragraphs from diverse sets of fiction books and magazines.

\paragraph{\indent {\small\faIcon{landmark}} government} The least represented genre, mainly denoting texts from governmental sources. These include political speeches (English-GUM by \citealp{en-gum}) as well as inscriptions from Neo-Assyrian kings from around 900 BCE (Akkadian-RIAO by \citealp{akk-riao}).

\paragraph{\indent {\small\faIcon{magic}} grammar-examples} Sentences from teaching or grammatical reference books which are typically short, but cover a wide range of dependency relations (e.g.\ Tagalog-TRG by \citealp{tl-trg}).

\paragraph{\indent {\small\faIcon{edit}} learner-essays} Small genre occurring in three single-genre treebanks. Sentences were written by second-language learners and either contain original errors (English-ESL by \citealp{en-esl}), manual corrections (IT-Valico by \citealp{it-valico}) or both (Chinese-CFL by \citealp{zh-cfl}).

\paragraph{\indent {\small\faIcon{gavel}} legal} Relatively frequent genre based mostly on laws and legal corpora within the public domain.

\paragraph{\indent {\small\faIcon{eye-dropper}} medical} Scientific articles/books in the field of medicine (e.g.\ cardiology, diabetes, endocrinology for Romanian-SiMoNERo by \citealp{ro-simonero}). It is subsumed by \textit{academic} for some treebanks (e.g.\ Czech-CAC by \citealp{cz-cac}).

\paragraph{\indent {\small\faIcon{newspaper}} news} The highest-resource genre by a large margin corresponding to news-wire texts as well as online newspapers on specific topics (e.g.\ IT-news in German-HDT by \citealp{de-hdt}).

\paragraph{\indent {\small\faIcon{info-circle}} nonfiction} Second most frequent genre with a high degree of variance, subsuming e.g.\ \textit{academic} and \textit{legal}. German-LIT~\cite{de-lit} contains three philosophical books from the 18th century. Other \textit{non-fiction} treebanks can originate from multiple sources (e.g.\ books and internet) and time spans.

\paragraph{\indent {\small\faIcon{music}} poetry} Smaller, yet distinct genre covering mostly older texts and language variations (e.g.\ Old French-SRCMF by \citealp{fro-srcmf}).

\paragraph{\indent {\small\faIcon{thumbs-up}} reviews} Medium-resource genre covering informal online reviews with unnormalized orthography (e.g.\ English-EWT) as well as formal reviews (e.g.\ newspaper film reviews in Czech-CAC).

\paragraph{\indent {\small\faIcon{rss}} social} Encompasses social media data such as tweets (e.g.\ Italian-TWITTIR{\`O} by \citealp{it-twittiro}) as well as newsgroups (e.g.\ English-EWT). Some \textit{spoken} data is co-labeled with this genre when it refers to colloquial speech (e.g. South Levantine Arabic-MADAR by \citealp{ajp-madar}).

\paragraph{\indent {\small\faIcon{comment}} spoken} Distinct genre which typically consists of spoken language transcriptions. Sentences contain filler words and may have abrupt boundaries. Sources range from elicited speech of native speakers (Komi Zyrian-IKDP by \citealp{kpv-tbs}) to radio program transcriptions (Frisian Dutch-Fame by \citealp{qfm-fame}).

\paragraph{\indent {\small\faIcon{globe}} web} Similarly ambiguous genre as \textit{non-fiction}. It occurs in conjunction with specific genres such as \textit{blog} and \textit{social} and never appears alone (e.g.\ Persian-PerDT by \citealp{fa-perdt}).

\paragraph{\indent {\small\faIcon{wikipedia-w}} wiki} Denotes data from Wikipedia for which cross-lingual authoring guidelines exist.

\begin{figure}
    \centering
    \hspace{-.5em}\pdftooltip{\includegraphics[width=.85\textwidth]{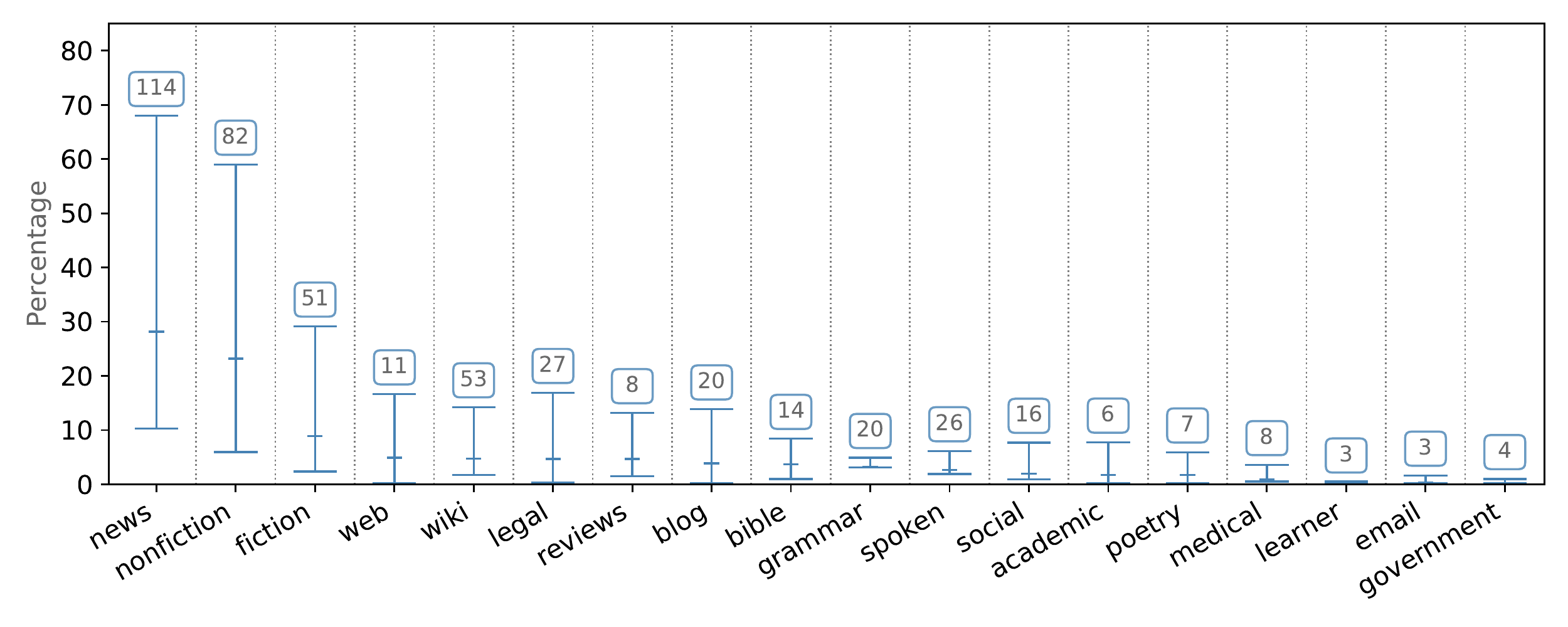}}{Screenreader Caption: news: part of 114 treebanks, 10\% to 68\% of sentences (28\% under uniformity). non-fiction: part of 82 treebanks, 6\% to 59\% of sentences (23\% under uniformity). fiction: part of 51 treebanks, 2\% to 29\% of sentences (9\% under uniformity). web: part of 11 treebanks, less than 1\% to 17\% of sentences (5\% under uniformity). wiki: part of 53 treebanks, 2\% to 14\% of sentences (5\% under uniformity). legal: part of 27 treebanks, less than 1\% to 17\% of sentences (5\% under uniformity). reviews: part of 8 treebanks, 1\% to 13\% of sentences (5\% under uniformity). blog: part of 20 treebanks, less than 1\% to 14\% of sentences (4\% under uniformity). bible: part of 14 treebanks, 1\% to 8\% of sentences (4\% under uniformity). grammar: part of 20 treebanks, 3\% to 5\% of sentences (3\% under uniformity). spoken: part of 26 treebanks, 2\% to 6\% of sentences (3\% under uniformity). social: part of 16 treebanks, 1\% to 8\% of sentences (2\% under uniformity). academic: part of 6 treebanks, less than 1\% to 8\% of sentences (2\% under uniformity). poetry: part of 7 treebanks, less than 1\% to 6\% of sentences (2\% under uniformity). medical: part of 8 treebanks, less than 1\% to 4\% of sentences (1\% under uniformity). learner: part of 3 treebanks, less than 1\% to less than 1\% of sentences (less than 1\% under uniformity). email: part of 3 treebanks, less than 1\% to 2\% of sentences (less than 1\% under uniformity). government: part of 4 treebanks, less than 1\% to 1\% of sentences (less than 1\% under uniformity).}
    \caption{\textbf{Genre Distribution in UD Version 2.8.} Ranges indicate upper/lower bounds for sentences per genre inferred from UD metadata. Center marker reflects the distribution under the assumption that genres within treebanks are uniformly distributed. Labels above the bars indicate the number of treebanks which contain each genre.}
    \label{fig:ud-genre-dist}
\end{figure}

\paragraph{} Figure \ref{fig:ud-genre-dist} shows the approximated distribution of these genres in UD. Maximum/minimum sentence counts are inferred from the size of single-genre treebanks plus the size of all treebanks in which a genre is said to occur. The center line denotes the distribution under the assumption that genres are uniformly distributed within each treebank.

It is clear that \textit{news} and \textit{non-fiction} constitute more than half of the entire dataset. Specialized genres such as \textit{medical} are less represented. For broader genres such as \textit{web}, which frequently co-occurs with others, the exact number of sentences is hard to estimate, but must lie between 0--20\%. Considering these large variances, access to instance-level genre will likely be crucial for effective proxy data selection and downstream domain adaptation.

\subsection{Instance-level Annotations}\label{sec:instance-annotations}

In addition to the aforementioned 18 treebank-level genre labels, some treebanks provide instance-level genre annotations in the comment-metadata before each sentence. We find such annotations in 26 out of 200 treebanks in UD 2.8 amounting to 124k or 8.25\% of all sentences.

Out of this set, 20 treebanks belong to the Parallel Universal Dependencies (PUD; \citealp{pud}). They are split 500/500 between \textit{news} and \textit{wiki}, as denoted by sentence IDs beginning with \texttt{n} and \texttt{w} respectively. The parallel nature of PUD makes it interesting for analyzing cross-lingual genre identification performance. However these two genres only represent a small fraction of non-fiction texts and furthermore, each PUD-treebank is test-split-only. Note also that Polish-PUD as an exception has the metadata labels \textit{news} and \textit{non-fiction}.

The remaining six treebanks for which we were able to identify instance-level genre annotations are Belarusian-HSE~\cite{be-hse}, Czech-CAC~\cite{cz-cac}, English-EWT~\cite{en-ewt}, German-LIT~\cite{de-lit}, Polish-LFG~\cite{pl-lfg} and Russian-Taiga~\cite{ru-taiga}. They cover a wider set of 12 genres. Annotation schema vary across treebanks and are neither fully compatible amongst each other nor with the 18 UD labels. Approximate mappings can however be drawn thanks to source data documentation by the respective authors (Section \ref{sec:supervised-eval}).

Further comment-metadata which may guide genre separation within treebanks includes document, paragraph and source identifiers. Again, these are unfortunately not available for all sentences (although coverage of these metadata reaches up to 45\%) and their values do not provide further indications about genre adherence.

\section{Instance Genre from Treebank Labels}\label{sec:methods}

From the previous analysis, it is evident that finer-grained genre labels are needed before domain adaptation can be successful across all languages.

Formally, the task of predicting instance-level UD genre can be defined as assigning a set of labels $\mathcal{L} = \{l_0, l_1, \ldots, l_K\}$ (i.e.\ genres) to all instances $x_n$ of a corpus $\mathcal{X}$ (i.e.\ UD). The corpus consists of $S$ distinct subsets $\mathcal{X} = \{\mathcal{X}_0 \cup \mathcal{X}_1 \cup \ldots \cup \mathcal{X}_S\}$ (i.e.\ treebanks) each with a subset of labels $\mathcal{L}_s \subseteq \mathcal{L}$. As no instance-level labels $x_n \rightarrow l$ are available, models must learn this mapping based solely on the subset of labels said to be contained in each data subset $\mathcal{X}_s \rightarrow \mathcal{L}_s$.

\subsection{Genre Prediction Methods}\label{sec:seg-methods}

As instance-level labels are noisy and sparse, we investigate two classification-based and two clustering-based approaches for inferring instance genre labels from the treebank metadata $\mathcal{L}_s$ alone. Building on \newcite{muellereberstein2021}, our proposed methods leverage latent genre information in the pre-trained mBERT language model~\cite{devlin2019}.

\paragraph{\indent \boot} In order to select proxy training data which matches the genre of an unseen target, \newcite{muellereberstein2021} propose a bootstrapping-based approach to genre classification (\boot{}).
An mBERT-based classifier~\cite{devlin2019} is initially trained on sentences from single-genre treebanks, corresponding to standard supervised classification. Above a confidence threshold (i.e.\ softmax probability of 0.99), sentences from treebanks containing a known genre in mixture are bootstrapped as single-genre training data for the next round. After bootstrapping sentences from all known genres, the remaining unclassified instances of any treebank containing a single unknown genre are inferred to be of that last genre.
While this method was previously used for targeted data selection, we investigate the degree to which it actually recovers instance-level genre.

\paragraph{\indent \class} With approximate classification (\class{}), we simplify \boot{} to naively learn instance genre labels from weak supervision. It fine-tunes the same mBERT MLM with a 18-genre classification layer on the \texttt{[CLS]}-token. For single-genre treebanks it is possible to measure the exact cross entropy between the predicted probability and the target (i.e.\ $x_n \rightarrow l$ with $l \in \mathcal{L}_s$ and $|\mathcal{L}_s| = 1$). For multi-genre treebanks with $|\mathcal{L}_s| > 1$, this is not possible as the gold label is unknown. For the \class{} approach, each sentence from a $k$-genre treebank is therefore classified $k$ times --- once for each class in $\mathcal{L}_s$.

\paragraph{\indent \gmm} In addition to classification, we also evaluate two common clustering algorithms. First we investigate whether clusters formed by untuned MLM sentence embeddings (mean over sentence sub-words) represent genre to such a degree that Gaussian Mixture Models can recover the 18 UD genre groups. For monolingual data from five genres, such clusters were shown to be recoverable~\cite{aharoni2020}. We extend this approach to the 114 language setting of UD.

\paragraph{\indent \lda} As all methods so far are to some degree dependent on the pre-trained MLM representations, we also evaluate the recoverability of genre using Latent Dirichlet Allocation~\cite{blei2003} with lexical features. Feature vectors are constructed using the frequency of character 3--6-grams.

\paragraph{\indent Cluster Labeling} Both clustering methods produce 18 groups of sentences from UD, however these will not carry meaningful labels as with classification. While labels could be assigned manually post-hoc by matching representative sentences in each cluster to one of the 18 global UD genres, this process is bound to be subjective and also depends on the annotator to be fluent in most of the 114 languages.

In order to automate this procedure, we propose \textbf{\gmml{}} and \textbf{\ldal{}} which combine clustering and classification. Both methods start by clustering each treebank $\mathcal{X}_s$ into the number of genres specified by its metadata (note that standard \gmm{} and \lda{} cluster all of UD at once, i.e.\ $\mathcal{X}$).

Next, the mean embedding of each cluster is computed such that they can be compared in a single representational space. Note that this would not be possible using monolingual models as their latent spaces are not as cross-lingually aligned. Analogous to \boot{}, single-genre treebanks can then be used as a single-label signal such that the closest cluster from each treebank containing the respective genre can be extracted. Newly identified clusters are added to the pool of single-genre clusters. This process need only be repeated for three rounds before all sentences in UD can be assigned a single label.

\paragraph{} Using these four methods, we aim to assign a single genre label to each sentence in UD. By comparing model ablations, we further depart from prior work and explicitly quantify the genre information in MLM embeddings as well as how it manifests within and across treebanks in UD.

\subsection{Supervised Evaluation}\label{sec:supervised-eval}

For the 26 treebanks with instance genre labels, we are able to measure standard \textsc{F1} after applying a mapping from the treebank-specific labels to the 18 global UD genre labels. The mapping was created according to the following criteria.

First, we only allowed treebank-specific genre labels to be mapped to the set of UD genre labels specified in each treebank's metadata.

Second, if possible treebank labels are mapped to UD labels of the same name (e.g.\ \texttt{fiction} $\rightarrow$ \textit{fiction}) or to the closest subsuming category (e.g.\ \texttt{spoken (prepared)} $\rightarrow$ \textit{spoken}).

Third, decisions involving subjective uncertainty were based on the label which covers the majority of data sources. E.g., Czech-CAC has the metadata label set \{\textit{legal, medical, news, non-fiction, reviews}\} and only three types of instance labels (\texttt{aw}, \texttt{nw}, \texttt{sw}). The \texttt{sw} (scientific-written) label is attached to many medical articles, but also to articles on philosophy or music. While \textit{academic} may be the most fitting label, it is not in the metadata. As such we chose the broader \textit{non-fiction} as the target label.

The full mapping is in Appendix \ref{sec:ud-setup} and we hope future work will be able to expand upon it.

\subsection{Unsupervised Evaluation}\label{sec:unsupervised-eval}

For the remaining 174 treebanks without sentence-level gold labels it is difficult to measure the exact quality of the predicted genre distributions. Nonetheless, treebank annotations provide enough information for approximate, global comparisons.

Based on label/cluster assignments, it is possible to compute the standard cluster purity measure (\textsc{Pur}; \citealp{schuetze2008}). Across treebanks of the same genre, the majority of sentences should belong to the same label/cluster. We measure this using the ratio of cross-treebank label agreement (\textsc{Agr}). As in prior work~\cite{aharoni2020} it is important to note that the aforementioned metrics can be misleading when taken on their own: A perfect score can for example be achieved by simply assigning all instances to the same genre.

To mitigate this issue we turn to the expected overlap of inter-treebank genre distributions. For multi-genre treebanks, it is known which genres are present, but not how they are distributed. Since treebanks are expected to have a certain amount of overlap, we can however estimate a global error. A \{\textit{fiction, spoken, wiki}\} treebank should for example have no clusters in common with a \{\textit{news}\} treebank, but should have many sentences in the same clusters as a \{\textit{fiction, medical, spoken}\} one. Assuming that genres are uniformly distributed within each treebank, the first pair would share 0 mass between distributions while the second pair would share $\frac{2}{3}$. Intuitively, a good prediction would produce a global genre distribution that falls precisely between the metadata range bars of Figure \ref{fig:ud-genre-dist}, close to the center markers.

To quantify the overlap between two treebank genre distributions $p$ and $q$ over the genres in $\mathcal{L}_s$, we use the discrete Bhattacharyya coefficient:
\begin{equation}
    \pdftooltip{BC(p,q) = \sum_{l \in \mathcal{L}_s} \sqrt{p(l) q(l)}}{Screenreader Caption: BC of p and q equals the sum of the square-root of p of l times q of l for all l in L subscript s.}
\end{equation}
which has often been applied to distributional comparisons ~\cite{choi2003,ruder2017}. It is computed for all pairs of treebanks such that the overlap error $\Delta$BC $\in [0, 100]$ is the mean absolute difference between the expected distributional overlap of each treebank pair and the predicted one (i.e.\ lower is better).
 
While none of these metrics can individually provide an exact measure of a prediction method's fit to the UD-specified distribution, they complement each other as to allow for global comparisons in absence of any sentence-level annotations.

\section{Experiments}\label{sec:experiments}

\subsection{Setup}\label{sec:setup}

\paragraph{Data} From the 1.5 million sentences in UD, we construct global training, development and testing splits. All original test splits are left unchanged and gathered into one global test split containing 204k sentences. Note that test-only treebanks and languages are thereby never seen during training or tuning. For instance-level, supervised evaluation, this means that all PUD treebanks and German-LIT are excluded, leaving five treebanks for tuning.

Next, all original training and development splits are concatenated and split 10/90 into a global training and development split with 102k and 915k sentences respectively. The reason for this small ``training'' split is that it is only required for training \class{} and \boot{}. Within it, we again split the data 70/30 (71k and 31k sentences) for classifier training and held-out data for early stopping. All exact splits are provided in Appendix \ref{sec:ud-setup}.

\paragraph{\indent Baselines} For our comparisons, we use a maximum frequency baseline (\freq{}) which labels all sentences within a treebank with the metadata genre label that is most frequent overall. For example, in any treebank containing \textit{news}, all instances are labeled as such.

In order to measure the untuned classification performance of mBERT, we propose an additional zero-shot classification baseline (\zero{}). Prior research has found that classifying sentences based solely on their cosine similarity to genre label strings in MLM embedding space can be remarkably effective~\cite{veeranna2016,yin2019,davison2020}. For example, a sentence is labeled as \textit{academic} if this is the closest embedded label out of all 18 genre strings.

\paragraph{\indent Training} Every method from Section \ref{sec:seg-methods} is run with three initializations. \class{} and \boot{} are trained for a maximum of 30 epochs with an early stopping patience of 3. \zero{}, \gmml{} and \ldal{} (by extension \gmm{}, \lda{}) do not require training and can be directly applied to the target data. Implementation details and development results are reported in Appendices \ref{sec:training-details} and \ref{sec:additional-results}.

\subsection{Results}\label{sec:results}

\begin{figure*}
    \centering
    \pdftooltip{\includegraphics[width=\textwidth]{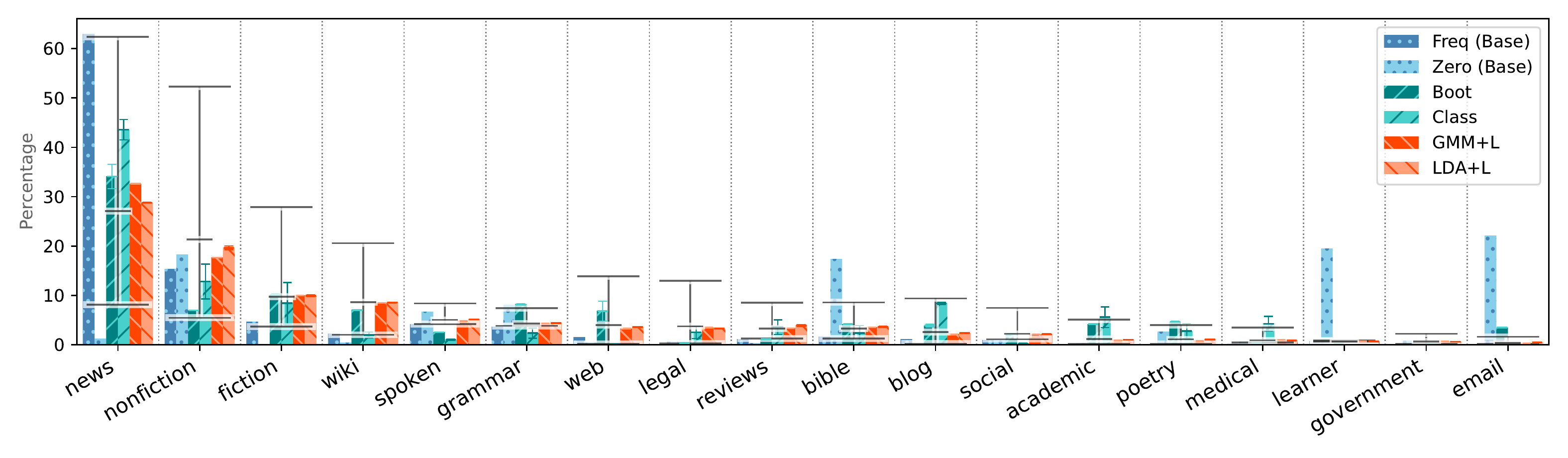}}{Screenreader Caption: news: UD metadata estimates 8\% to 62\% of sentences (27\% under uniformity). Freq 63\%, Zero 1\%, Boot 34\%, Class 44\%, GMM+L 33\%, LDA+L 29\%. non-fiction: UD metadata estimates 5\% to 52\% of sentences (21\% under uniformity). Freq 15\%, Zero 18\%, Boot 7\%, Class 13\%, GMM+L 18\%, LDA+L 20\%. fiction: UD metadata estimates 3\% to 28\% of sentences (10\% under uniformity). Freq 5\%, Zero less than 1\%, Boot 10\%, Class 8\%, GMM+L 10\%, LDA+L 10\%. wiki: UD metadata estimates 2\% to 21\% of sentences (9\% under uniformity). Freq 2\%, Zero less than 1\%, Boot 7\%, Class 2\%, GMM+L 8\%, LDA+L 9\%. spoken: UD metadata estimates 4\% to 8\% of sentences (5\% under uniformity). Freq 4\%, Zero 7\%, Boot 3\%, Class 1\%, GMM+L 5\%, LDA+L 5\%. grammar: UD metadata estimates 4\% to 7\% of sentences (4\% under uniformity). Freq 4\%, Zero 8\%, Boot 8\%, Class 2\%, GMM+L 4\%, LDA+L 4\%. web: UD metadata estimates less than 1\% to 14\% of sentences (4\% under uniformity). Freq 1\%, Zero less than 1\%, Boot 7\%, Class less than 1\%, GMM+L 3\%, LDA+L 4\%. legal: UD metadata estimates less than 1\% to 13\% of sentences (4\% under uniformity). Freq less than 1\%, Zero 1\%, Boot less than 1\%, Class 3\%, GMM+L 3\%, LDA+L 3\%. reviews: UD metadata estimates 1\% to 9\% of sentences (3\% under uniformity). Freq 1\%, Zero less than 1\%, Boot 1\%, Class 4\%, GMM+L 3\%, LDA+L 4\%. bible: UD metadata estimates 1\% to 9\% of sentences (3\% under uniformity). Freq 2\%, Zero 17\%, Boot 4\%, Class 2\%, GMM+L 4\%, LDA+L 4\%. blog: UD metadata estimates less than 1\% to 9\% of sentences (3\% under uniformity). Freq 1\%, Zero less than 1\%, Boot 4\%, Class 8\%, GMM+L 2\%, LDA+L 2\%. social: UD metadata estimates 1\% to 7\% of sentences (2\% under uniformity). Freq 1\%, Zero 1\%, Boot 2\%, Class less than 1\%, GMM+L 2\%, LDA+L 2\%. academic: UD metadata estimates less than 1\% to 5\% of sentences (1\% under uniformity). Freq less than 1\%, Zero less than 1\%, Boot 4\%, Class 6\%, GMM+L 1\%, LDA+L 1\%. poetry: UD metadata estimates less than 1\% to 4\% of sentences (1\% under uniformity). Freq less than 1\%, Zero 3\%, Boot 5\%, Class 3\%, GMM+L 1\%, LDA+L 1\%. medical: UD metadata estimates less than 1\% to 3\% of sentences (1\% under uniformity). Freq less than 1\%, Zero 1\%, Boot less than 1\%, Class 4\%, GMM+L 1\%, LDA+L 1\%. learner: UD metadata estimates 1\% to 1\% of sentences (1\% under uniformity). Freq 1\%, Zero 19\%, Boot less than 1\%, Class less than 1\%, GMM+L 1\%, LDA+L 1\%. government: UD metadata estimates less than 1\% to 2\% of sentences (1\% under uniformity). Freq less than 1\%, Zero 1\%, Boot less than 1\%, Class less than 1\%, GMM+L 1\%, LDA+L 1\%. email: UD metadata estimates less than 1\% to 2\% of sentences (less than 1\% under uniformity). Freq less than 1\%, Zero 22\%, Boot 3\%, Class less than 1\%, GMM+L less than 1\%, LDA+L less than 1\%.}
    \caption{\textbf{Genre Predictions on UD (Test).} Ranges indicate upper/lower bounds inferred from UD metadata and the distribution under treebank-level uniformity at the center marker. Bars show averaged distribution predictions with standard deviations by \freq{}, \zero{}, \boot{}, \class{}, \gmml{} and \ldal{}.}
    \label{fig:pred-distributions}
\end{figure*}

\begin{table}
\centering
\resizebox{.48\textwidth}{!}{
\begin{tabular}{lrrrr}
\toprule
\textsc{Method} & \textsc{Pur} & \textsc{Agr} & \textsc{$\Delta$BC} & \multicolumn{1}{|r}{\textsc{F1}} \\
\midrule
\freq & 100{\footnotesize$\pm$0.0} & 100{\footnotesize$\pm$0.0} & 21{\footnotesize$\pm$0.0} & \multicolumn{1}{|r}{47{\footnotesize$\pm$0.0}} \\
\zero & 46{\footnotesize$\pm$0.0} & 56{\footnotesize$\pm$0.0} & 47{\footnotesize$\pm$0.0} & \multicolumn{1}{|r}{12{\footnotesize$\pm$0.0}} \\
\midrule
\class & 83{\footnotesize$\pm$1.4} & 63{\footnotesize$\pm$3.9} & 34{\footnotesize$\pm$1.1} & \multicolumn{1}{|r}{32{\footnotesize$\pm$0.9}} \\ 
\boot & 86{\footnotesize$\pm$0.4} & 70{\footnotesize$\pm$0.7} & 29{\footnotesize$\pm$0.3} & \multicolumn{1}{|r}{38{\footnotesize$\pm$1.2}} \\ 
\midrule
\gmm & 90{\footnotesize$\pm$0.5} & 45{\footnotesize$\pm$2.6} & 31{\footnotesize$\pm$0.3} & \multicolumn{1}{|r}{---}\\ 
\footnotesize{\textsc{+Labels}} & 100{\footnotesize$\pm$0.0} & 100{\footnotesize$\pm$0.0} & 4{\footnotesize$\pm$0.2} & \multicolumn{1}{|r}{54{\footnotesize$\pm$2.1}} \\ 
\lda & 77{\footnotesize$\pm$0.8} & 34{\footnotesize$\pm$2.6} & \multicolumn{1}{r}{31{\footnotesize$\pm$0.2}} & \multicolumn{1}{|r}{---} \\ 
\footnotesize{\textsc{+Labels}} & 100{\footnotesize$\pm$0.0} & 100{\footnotesize$\pm$0.0} & 2{\footnotesize$\pm$0.1} & \multicolumn{1}{|r}{51{\footnotesize$\pm$1.5}} \\ 
\bottomrule
\end{tabular}}
\caption{\label{tab:results-global} \textbf{Results of Genre Prediction on UD (Test).} Purity (\textsc{Pur} $\uparrow$), agreement (\textsc{Agr} $\uparrow$), overlap error ($\Delta$BC $\downarrow$) and micro-F1 over instance-labeled TBs (\textsc{F1} $\uparrow$) for \freq{}, \zero{}, \class{}, \boot{} and \gmm{}, \lda{} with/without cluster label predictions (\textsc{+Labels}). Standard deviation denoted $\pm$.}
\end{table}

Using the 8\% subset of annotated instances (Section \ref{sec:supervised-eval}) in addition to the unsupervised metrics from Section \ref{sec:unsupervised-eval}, we can gather an estimate of each method's performance in Table \ref{tab:results-global}. UD-level genre predictions in addition to instance-level confusions are further visualized in Figures \ref{fig:pred-distributions} and \ref{fig:confusions}.

\paragraph{\indent Baselines} The \freq{} baseline highlights the issue of using individual unsupervised metrics for estimating performance. As it assigns all sentences per treebank to the same genre, it automatically achieves 100\% single-genre treebank purity and agreement. Considering that the instance-level F1 covers 12 genres, a baseline score of 47 is also competitive. Note that this is mostly due to the data imbalance towards \textit{news}. This unlikely distribution predicted by \freq{} is also reflected in Figure \ref{fig:pred-distributions}.

\zero{}-shot classification is not fine-tuned on UD-specific signals and as such predicts a genre distribution that does not adhere to the metadata at all (see Figure \ref{fig:pred-distributions}). It severely underpredicts high-frequency genres such as \textit{news} and overpredicts less frequent genres such as \textit{email}. This reflects in our metrics, with \zero{} obtaining the lowest \textsc{Pur}, \textsc{Agr} and F1 while having the highest \textsc{$\Delta$BC} of 47.

\paragraph{\indent Classification} With regard to explicit genre fine-tuning, \class{} increases purity by 38 points compared to \zero{}. Agreement across treebanks also improves, while overlap error decreases. These differences are also reflected in Figure \ref{fig:pred-distributions} in that the predicted distribution is more within the range that would be expected given the metadata.

\boot{} fits the UD genre distribution more closely, resulting in a purity that is 4 points higher and agreement that is 11 points higher than \class{}. F1 also increases by 6 points while overlap error decreases by 4 points, indicating that these improvements are not merely due to e.g.\ assigning all sentences to the same genre. While instance-level F1 is below the \freq{} baseline, both methods improve upon the untuned \zero{} by a factor of 3.

The benefits of the less noisy training signal are visible in Figure \ref{fig:pred-distributions}: Compared to \class{}, \boot{} predicts labels in a way that more closely resembles the expected distribution even when the label only occurs in multi-genre treebanks and is ambiguous (e.g.\ \textit{web}). While \boot{} agrees upon the same genre-label across languages (e.g.\ all \textit{social} treebanks are labeled as such), \class{} tends to overassign the globally most frequent labels (e.g.\ half of \textit{social} treebanks are labeled \textit{wiki}) and has a larger variance in its assignments across initializations.

\paragraph{\indent Clustering} \gmm{} clusters from untuned mBERT embeddings follow the distribution specified by UD metadata more than the \lda{} clusters produced from lexical information. Although sentence representations are gathered using a naive mean-pooling approach, the resulting clusters reach 90\% \textsc{Pur} compared to 77\% for \lda{}. \textsc{Agr} follows a similar pattern and \textsc{$\Delta$BC} is equivalent.

Turning to our cluster labelling approaches, both \gmml{} and \ldal{} obtain the highest overall F1 scores, outperforming both baselines. They achieve 100\% \textsc{Pur} and \textsc{Agr} by the same process as the \freq{}-baseline while their overlap error is significantly lower at 4 and 2 points respectively. Figure \ref{fig:pred-distributions} reflects this, as \gmml{} and \ldal{} are always closest to the expected genre distribution, regardless of overall genre frequency. This shows how focusing on treebank-internal differences before applying a global labelling procedure combines the benefits of local clustering with the benefits of bootstrapped classification, resulting in an effective overall method.

\subsection{Analysis}\label{sec:analysis}

\begin{figure}
    \centering
    \begin{subfigure}[b]{.30\textwidth}
        \raisebox{.15cm}{\pdftooltip{\includegraphics[height=5cm]{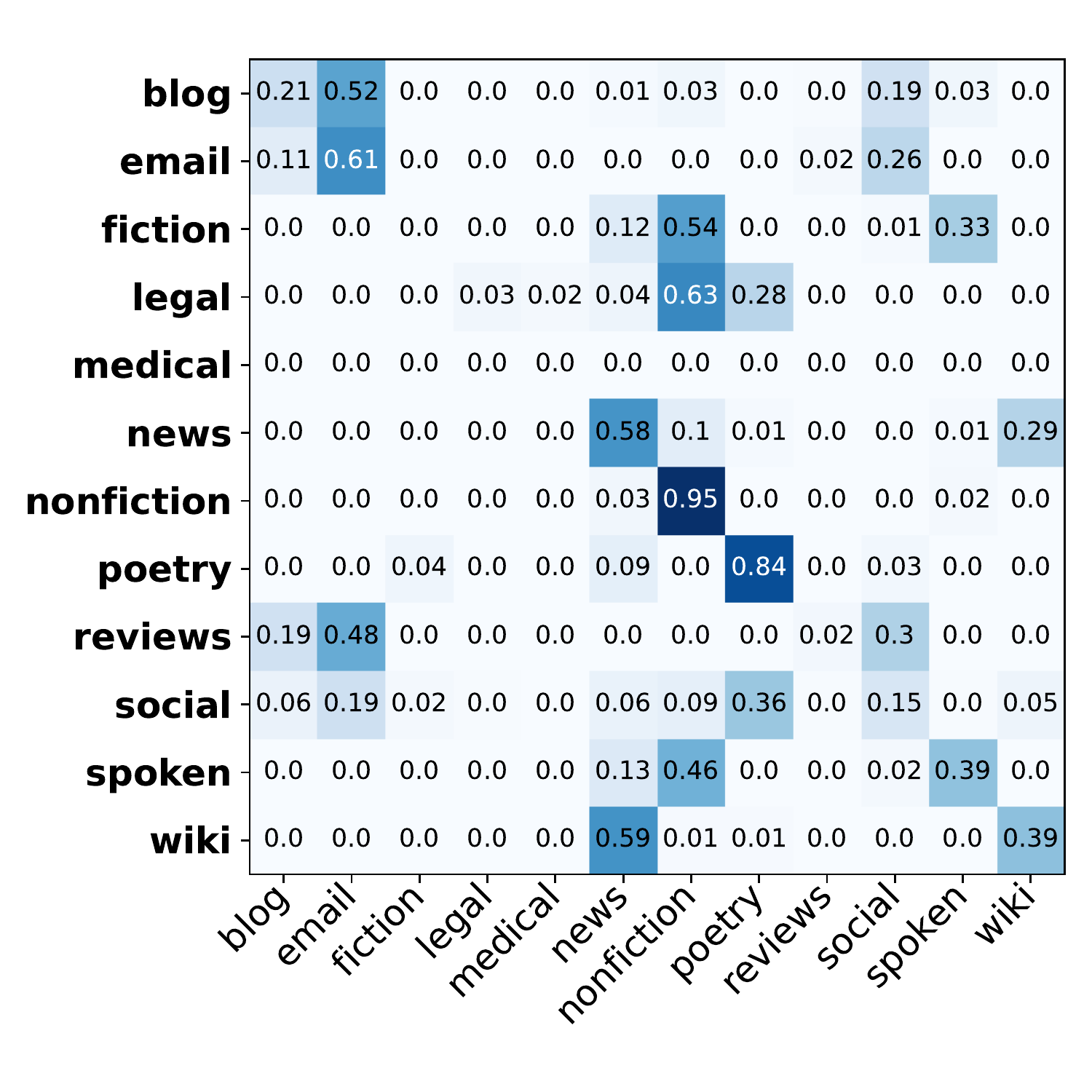}}{Screenreader Caption: blog (target): 21\% blog, 52\% email, 1\% news, 3\% non-fiction, 19\% social, 3\% spoken. email (target): 11\% blog, 61\% email, 2\% reviews, 26\% social. fiction (target): 12\% news, 54\% non-fiction, 1\% social, 33\% spoken. legal (target): 3\% legal, 2\% medical, 4\% news, 63\% non-fiction, 28\% poetry. medical (target). news (target): 58\% news, 10\% non-fiction, 1\% poetry, 1\% spoken, 29\% wiki. non-fiction (target): 3\% news, 95\% non-fiction, 2\% spoken. poetry (target): 4\% fiction, 9\% news, 84\% poetry, 3\% social. reviews (target): 19\% blog, 48\% email, 2\% reviews, 30\% social. social (target): 6\% blog, 19\% email, 2\% fiction, 6\% news, 9\% non-fiction, 36\% poetry, 15\% social, 5\% wiki. spoken (target): 13\% news, 46\% non-fiction, 2\% social, 39\% spoken. wiki (target): 59\% news, 1\% non-fiction, 1\% poetry, 39\% wiki.}}
        \vspace{-.9cm}
        \caption{\zero{}}
        \label{fig:confusions-zero}
    \end{subfigure}
    \begin{subfigure}[b]{.30\textwidth}
        \raisebox{.35cm}{\pdftooltip{\includegraphics[height=4.65cm]{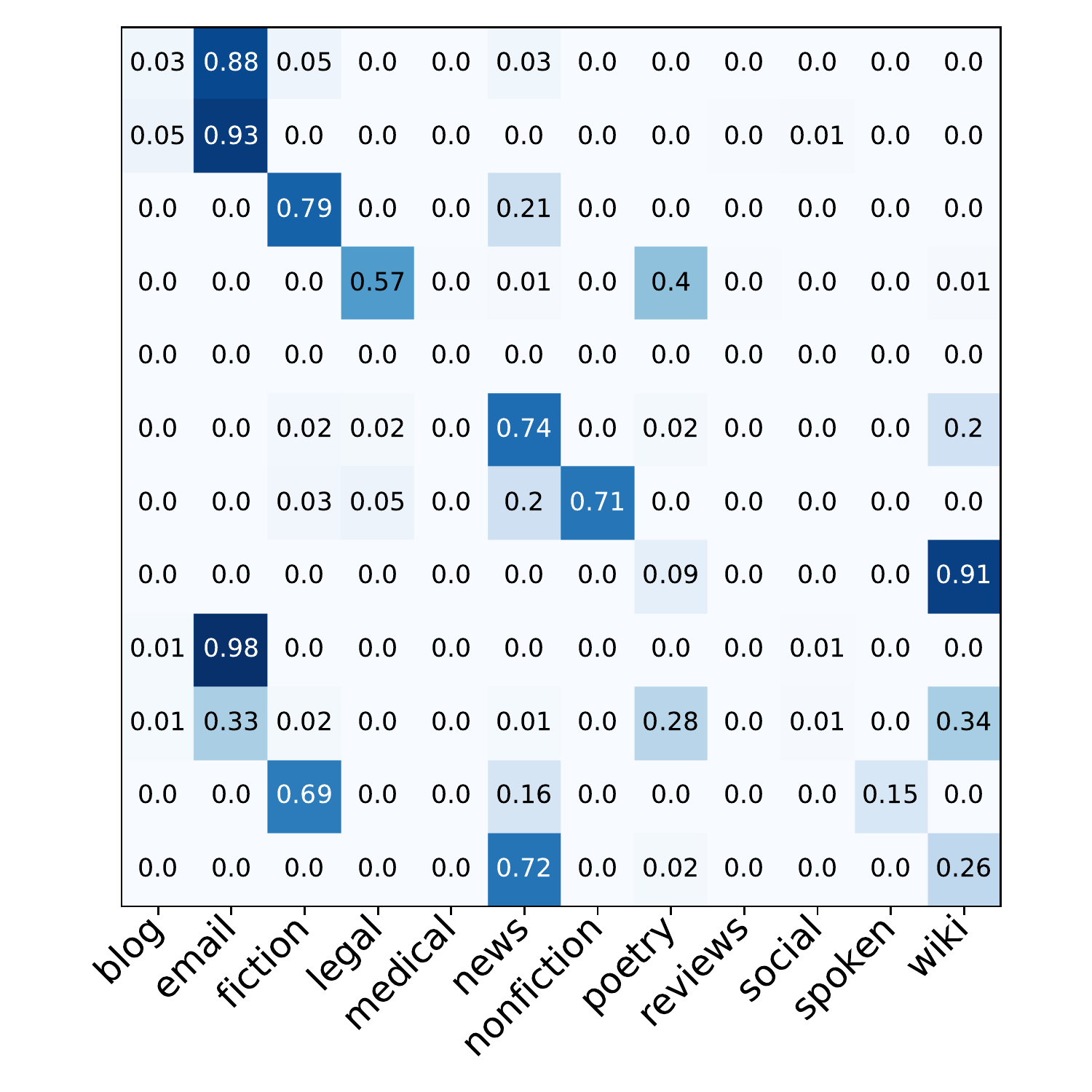}}{Screenreader Caption: blog (target): 3\% blog, 88\% email, 5\% fiction, 3\% news. email (target): 5\% blog, 93\% email, 1\% social. fiction (target): 79\% fiction, 21\% news. legal (target): 57\% legal, 1\% news, 40\% poetry, 1\% wiki. medical (target). news (target): 2\% fiction, 2\% legal, 74\% news, 2\% poetry, 20\% wiki. non-fiction (target): 3\% fiction, 5\% legal, 20\% news, 71\% non-fiction. poetry (target): 9\% poetry, 91\% wiki. reviews (target): 1\% blog, 98\% email, 1\% social. social (target): 1\% blog, 33\% email, 2\% fiction, 1\% news, 28\% poetry, 1\% social, 34\% wiki. spoken (target): 69\% fiction, 16\% news, 15\% spoken. wiki (target): 72\% news, 2\% poetry, 26\% wiki.}}
        \vspace{-.45cm}
        \caption{\boot{}}
        \label{fig:confusions-boot}
    \end{subfigure}
    \hspace{-.5cm}
    \begin{subfigure}[b]{.30\textwidth}
        \pdftooltip{\includegraphics[height=5.2cm]{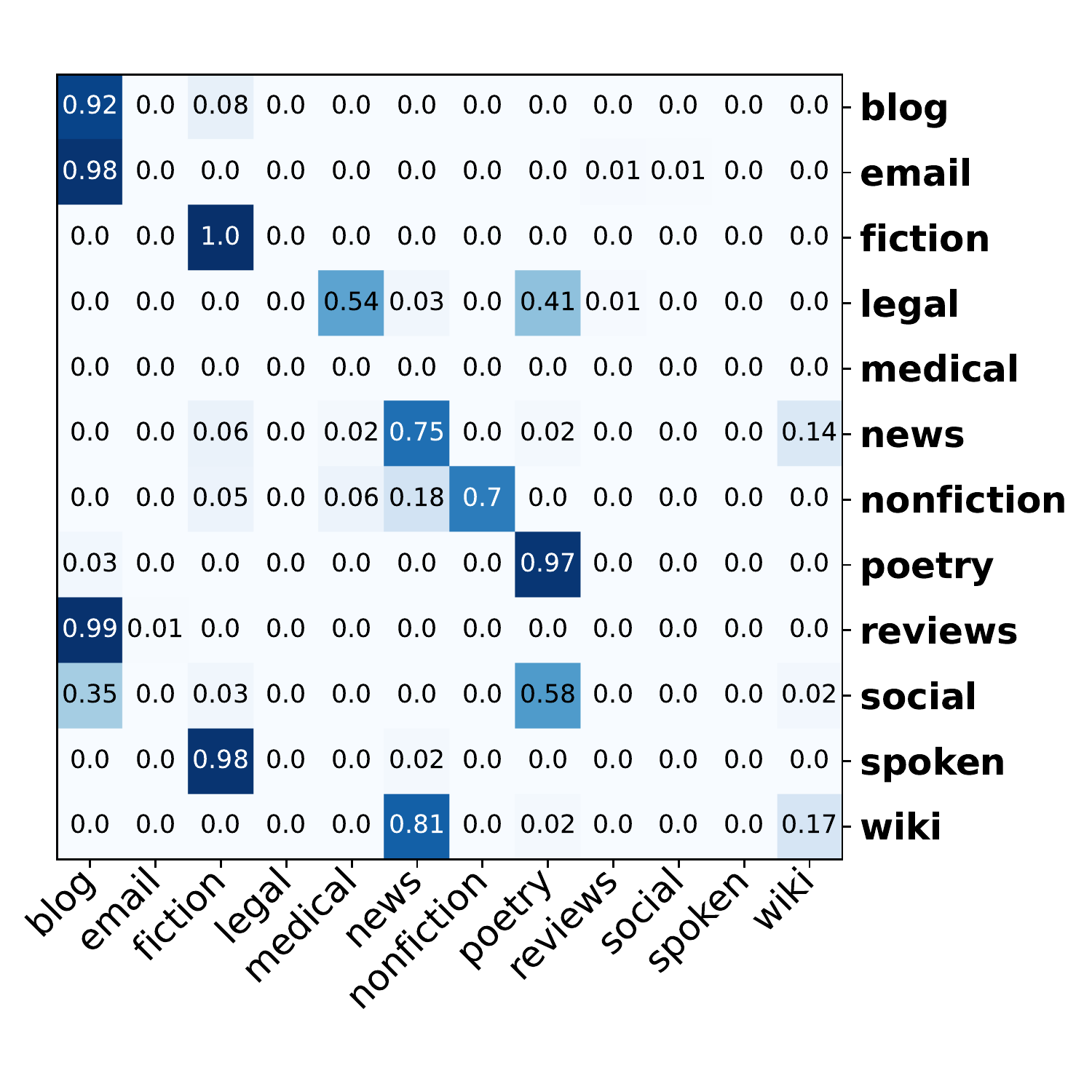}}{Screenreader Caption: blog (target): 92\% blog, 8\% fiction. email (target): 98\% blog, 1\% reviews, 1\% social. fiction (target): 100\% fiction. legal (target): 54\% medical, 3\% news, 41\% poetry, 1\% reviews. medical (target). news (target): 6\% fiction, 2\% medical, 75\% news, 2\% poetry, 14\% wiki. non-fiction (target): 5\% fiction, 6\% medical, 18\% news, 70\% non-fiction. poetry (target): 3\% blog, 97\% poetry. reviews (target): 99\% blog, 1\% email. social (target): 35\% blog, 3\% fiction, 58\% poetry, 2\% wiki. spoken (target): 98\% fiction, 2\% news. wiki (target): 81\% news, 2\% poetry, 17\% wiki.}
        \vspace{-.9cm}
        \caption{\class{}}
        \label{fig:confusions-class}
    \end{subfigure}
    \hspace{-.75em}
    \begin{subfigure}[b]{.30\textwidth}
        \raisebox{.15cm}{\pdftooltip{\includegraphics[height=5cm]{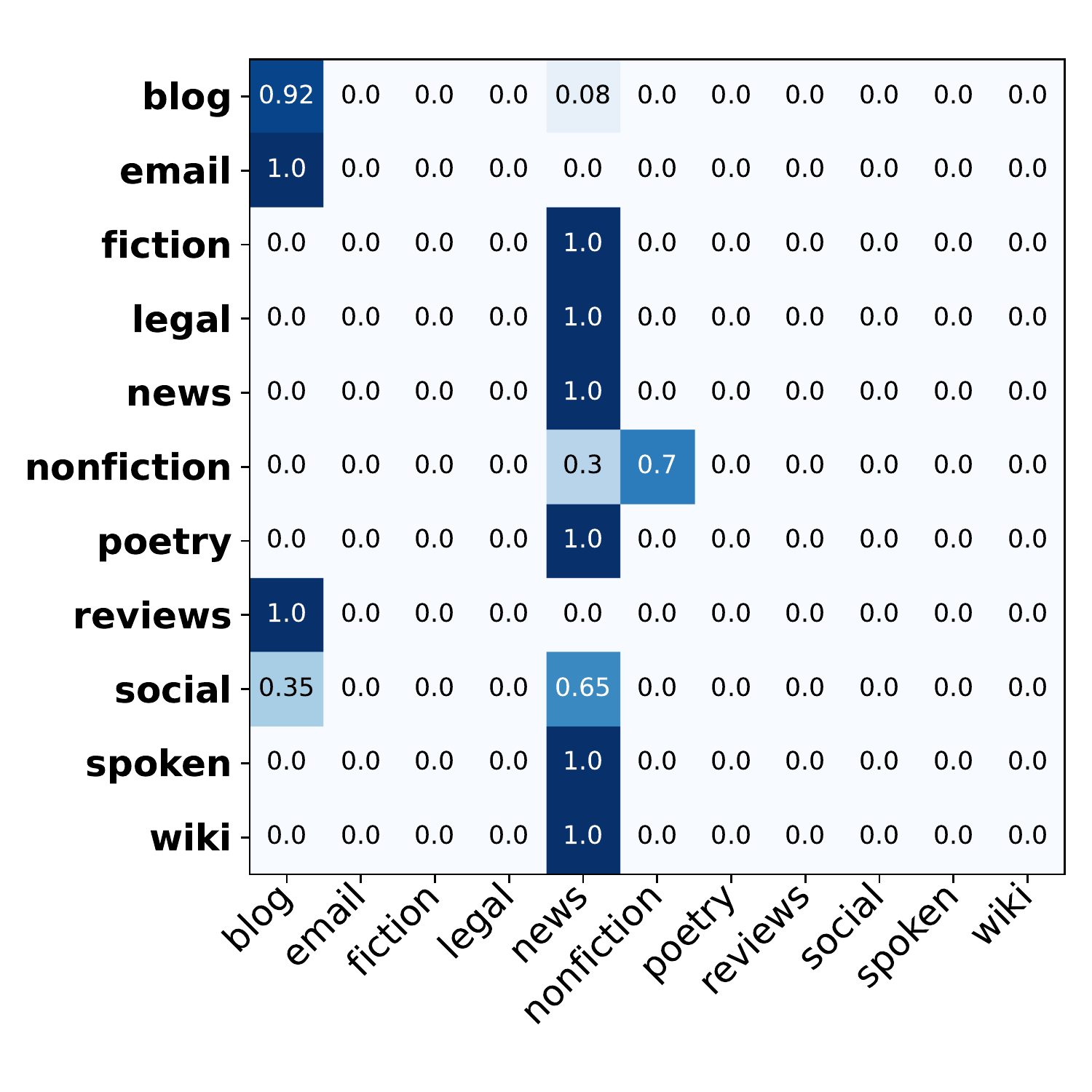}}{Screenreader Caption: blog (target): 92\% blog, 8\% news. email (target): 100\% blog. fiction (target): 100\% news. legal (target): 100\% news. news (target): 100\% news. non-fiction (target): 30\% news, 70\% non-fiction. poetry (target): 100\% news. reviews (target): 100\% blog. social (target): 35\% blog, 65\% news. spoken (target): 100\% news. wiki (target): 100\% news.}}
        \vspace{-.9cm}
        \caption{\textsc{Freq}}
        \label{fig:confusions-freq}
    \end{subfigure}
    \begin{subfigure}[b]{.30\textwidth}
        \raisebox{.35cm}{\pdftooltip{\includegraphics[height=4.65cm]{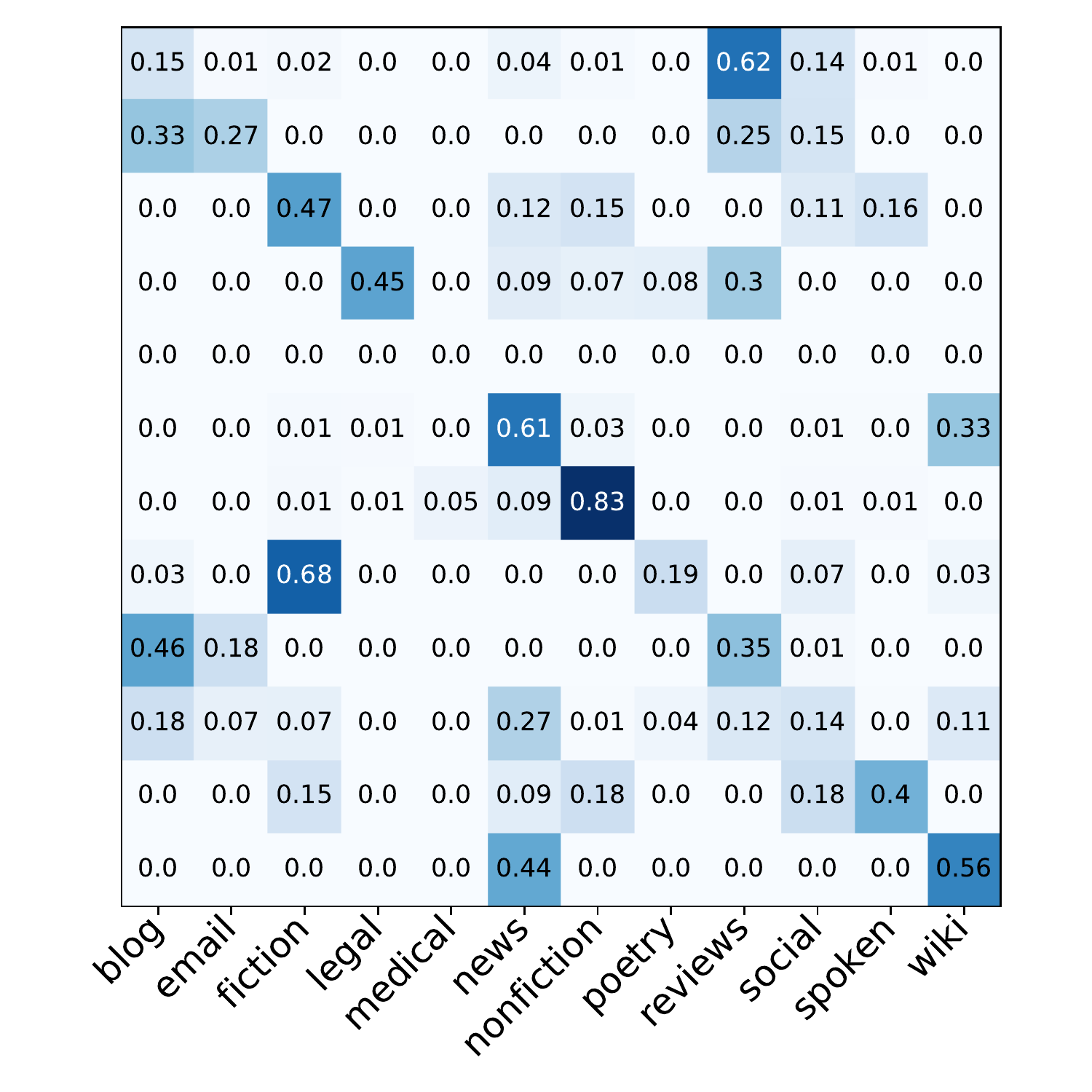}}{Screenreader Caption: blog (target): 15\% blog, 1\% email, 2\% fiction, 4\% news, 1\% non-fiction, 62\% reviews, 14\% social, 1\% spoken. email (target): 33\% blog, 27\% email, 25\% reviews, 15\% social. fiction (target): 47\% fiction, 12\% news, 15\% non-fiction, 11\% social, 16\% spoken. legal (target): 45\% legal, 9\% news, 7\% non-fiction, 8\% poetry, 30\% reviews. medical (target). news (target): 1\% fiction, 1\% legal, 61\% news, 3\% non-fiction, 1\% social, 33\% wiki. non-fiction (target): 1\% fiction, 1\% legal, 5\% medical, 9\% news, 83\% non-fiction, 1\% social, 1\% spoken. poetry (target): 3\% blog, 68\% fiction, 19\% poetry, 7\% social, 3\% wiki. reviews (target): 46\% blog, 18\% email, 35\% reviews, 1\% social. social (target): 18\% blog, 7\% email, 7\% fiction, 27\% news, 1\% non-fiction, 4\% poetry, 12\% reviews, 14\% social, 11\% wiki. spoken (target): 15\% fiction, 9\% news, 18\% non-fiction, 18\% social, 40\% spoken. wiki (target): 44\% news, 56\% wiki.}}
        \vspace{-.45cm}
        \caption{\gmml{}}
        \label{fig:confusions-gmml}
    \end{subfigure}
    \hspace{-.5cm}
    \begin{subfigure}[b]{.30\textwidth}
        \pdftooltip{\includegraphics[height=5.2cm]{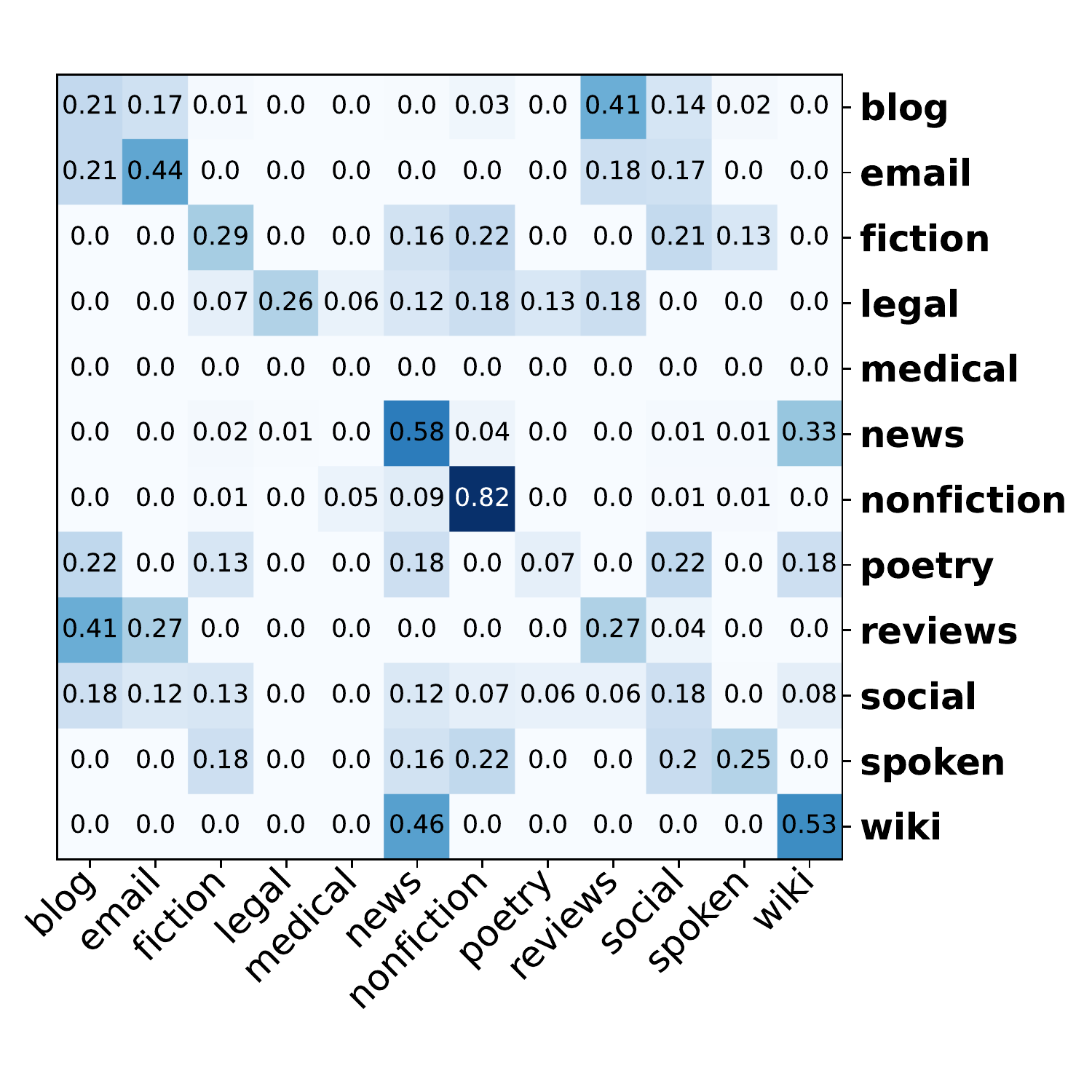}}{Screenreader Caption: blog (target): 21\% blog, 17\% email, 1\% fiction, 3\% non-fiction, 41\% reviews, 14\% social, 2\% spoken. email (target): 21\% blog, 44\% email, 18\% reviews, 17\% social. fiction (target): 29\% fiction, 16\% news, 22\% non-fiction, 21\% social, 13\% spoken. legal (target): 7\% fiction, 26\% legal, 6\% medical, 12\% news, 18\% non-fiction, 13\% poetry, 18\% reviews. medical (target). news (target): 2\% fiction, 1\% legal, 58\% news, 4\% non-fiction, 1\% social, 1\% spoken, 33\% wiki. non-fiction (target): 1\% fiction, 5\% medical, 9\% news, 82\% non-fiction, 1\% social, 1\% spoken. poetry (target): 22\% blog, 13\% fiction, 18\% news, 7\% poetry, 22\% social, 18\% wiki. reviews (target): 41\% blog, 27\% email, 27\% reviews, 4\% social. social (target): 18\% blog, 12\% email, 13\% fiction, 12\% news, 7\% non-fiction, 6\% poetry, 6\% reviews, 18\% social, 8\% wiki. spoken (target): 18\% fiction, 16\% news, 22\% non-fiction, 20\% social, 25\% spoken. wiki (target): 46\% news, 53\% wiki.}
        \vspace{-.9cm}
        \caption{\ldal{}}
        \label{fig:confusions-ldal}
    \end{subfigure}
    \caption{\textbf{Confusions of Instance-level Genre.} Ratios of predicted labels (columns) per target (row) for \zero{}, \boot{}, \class{}, \freq{}, \gmml{}, \ldal{} on test splits of 26 instance-annotated treebanks.}
    \label{fig:confusions}
\end{figure}

From the F1 scores in Table \ref{tab:results-global} it is clear that predicting instance genre based on treebank metadata alone --- while accounting for its skewed distribution and inter-treebank shifts of genre definitions --- is a difficult task. In the following we analyze the performance characteristics of each method.

Overall, trends of the unsupervised metrics follow the supervised F1, leading us to believe that the methods would behave comparatively should labels for all instances in UD be available. The confusion matrices with prediction ratios per gold label in Figure \ref{fig:confusions} reflect our previous observations.

\paragraph{\indent Baselines} The \freq{} baseline's predictions are clearly dominated by the most frequent \textit{news} genre, followed by the similarly high frequency \textit{non-fiction} and \textit{blog} (see Figure \ref{fig:confusions-freq}).

\zero{} appears to follow a pattern similar to \boot{} (e.g.\ \textit{blog} and \textit{email}), however it also makes more predictions away from the diagonal (see Figure \ref{fig:confusions-zero}).

\paragraph{\indent Classification}  Both \class{} (Figure \ref{fig:confusions-class}) and \boot{} (Figure \ref{fig:confusions-boot}) assign most instances of a genre to a single prediction label, often strongly aligning with the target diagonal. \class{} more often assigns a single label per target instead of spreading out predictions across multiple labels as in \boot{}. Nonetheless, both methods make some unintuitive errors such as \boot{} classifying parts of \textit{poetry} as \textit{wiki}. For these 68 samples from Russian-Taiga, \boot{} likely overfits the language signal from Russian-GSD (\citealp{ru-gsd}; \textit{wiki}).

Compared to \zero{} which approximates the predictions of an untuned mBERT model, \boot{} and \class{} fine-tuning appears to amplify existing patterns and shifts some predictions to better align with genres as defined in UD (e.g.\ \textit{fiction} and \textit{legal} in \boot{}).

\paragraph{\indent Clustering} Grouping all 1.5 million sentences of UD into 18 unlabeled clusters using \gmm{} and \lda{} results in purity and \textsc{$\Delta$BC} comparable to \class{} and \boot{}. However, looking into the cluster contents of the former reveals that they are oversaturated with large treebanks such as German-HDT. Cosine similarities of cluster centroids from the mBERT-based \gmm{} further indicate that proximity corresponds foremost to language similarity.

Some clusters predominantly contain \textit{news}, \textit{wiki} or \textit{social}. This corresponds to cases such as the Italian Twitter treebank TWITTIR{\`O} in which specific tokens (e.g.\ ``@user'') are distinct enough to override the language signal. Overall, most UD-level clusters do not have clear genre distinctions and are influenced more strongly by language than genre, resulting in high treebank purity while having low intra-treebank agreement. Attempting to cross-lingually cluster all sentences in UD directly is therefore not as effective for recovering instance-level genre as it was in the monolingual setting~\cite{aharoni2020}.

Initially constructing clusters within each treebank as in the \gmml{} and \ldal{} methods appears to restore the benefits observed in the monolingual setting. A qualitative analysis of the treebank-level LDA clusters reveals that \textit{wiki} clusters often contain lexical indicators for the genre, such as brackets, while \textit{news} features often contain n-grams which may be related to spoken quotes such as ``said'', ``Ik\textvisiblespace'' (first person pronoun).

Attaching labels to these clusters using the globally shared mBERT space yields confusion plots for \gmml{} and \ldal{} which most closely follow the diagonal (see Figures \ref{fig:confusions-gmml} and \ref{fig:confusions-ldal}). Overall, their predictions follow a similar pattern indicating that clustering at the treebank-level using either mBERT embeddings or lexical features results in similar sentence groups.

Within the instance-labeled subset, all models share confusions between \textit{news} and \textit{wiki} (mainly from PUD). While \textit{wiki} is often predicted as \textit{news}, both \gmml{} and \ldal{} substantially improve upon this ``\textit{news}-bias'' with a confusion ratio that is 13\%--56\% lower compared to all other methods. The sentence-bounded context from which all models must make their genre predictions nonetheless limits the amount of improvement possible. For example, using the aforementioned LDA features the algorithm would very likely be unable to distinguish between \textit{news} and \textit{wiki} (both non-fiction, edited texts describing facts) for cases such as, \textit{``Weiss was honored with the literature prizes from the cities of Cologne and Bremen.''}

\section{Discussion and Conclusion}

This work provided an in-depth analysis of the 18 genres in Universal Dependencies (UD) and identified challenges for projecting this treebank metadata to the instance level. As these genre labels were not part of the first UD releases, but were added in later versions, we identified large variations in the way they are interpreted and applied --- resulting in far less universal definitions of genre than for syntactic dependencies. Most treebanks furthermore contain multiple genres while not providing finer-grained instance-level annotations thereof. This also sheds light on prior work which used UD metadata for training data selection, where treebank-level genre improved in-language parsing performance~\cite{stymne-2020-cross} and where moving to instance-level genre signals lead to additional increases even across languages~\cite{muellereberstein2021}.

Building on the latent genre information stored in MLM embeddings, we investigated four methods for projecting treebank-level labels to the instance level. In contrast to prior monolingual work, immediately clustering multilingual embeddings yielded clusters dominated by language similarity instead of genre (Section \ref{sec:analysis}). Similarly, zero-shot labelling using the untuned mBERT latent space proved to be insufficient for producing a genre distribution which adheres to the UD metadata. The classification-based \class{} and \boot{} methods are able to extract a stronger genre signal from mBERT than \zero{}.

Our proposed \gmml{} and \ldal{} methods which combine local treebank clusters with the global, cross-lingual representation space reach the best overall performance, outperforming both baselines as well as both classification methods at a much lower computational cost (Section \ref{sec:results}; Appendix \ref{sec:training-details}). This highlights how the current genre annotations are far from universal, yet can still guide our local-to-global instance-level genre predictors in identifying cross-lingually consistent, data-driven notions of genre.

Future work may be able to improve instance genre prediction by using a more consistent label set or human annotations. The definition of genre macro-classes or a broader taxonomy covering existing annotations could also guide further investigations into cross-lingual language variation. Nonetheless, we expect the task of predicting sentence genre to remain difficult due to the short context within which both annotators and models must make their predictions.

Within the complex scenario of highly cross-lingual, instance-level genre classification, our methods have nonetheless demonstrated that genre is recoverable across the 114 languages in UD --- shedding light on prior genre-driven work as well as enabling future research to more deliberately control for additional dimensions of language variation in their data.

\section*{Acknowledgements}

We would like to thank the NLPnorth group for insightful discussions on this work --- in particular Elisa Bassignana and Mike Zhang. Thanks to Héctor Martínez Alonso for feedback on an early draft as well as ITU's High-performance Computing Cluster team. Finally, we thank the anonymous reviewers for their helpful feedback. This research is supported by the Independent Research Fund Denmark (DFF) grant 9063-00077B and an Amazon Faculty Research Award (ARA).

\bibliographystyle{acl}
\bibliography{anthology,coling2020,references}

%
%

\appendix

\section*{Appendix}

\section{Universal Dependencies Setup}\label{sec:ud-setup}
All experiments make use of Universal Dependencies v2.8~\cite{ud28}. From the total set of 202 treebanks, we use all except for the following two (due to licensing restrictions): \textit{Arabic-NYUAD} and \textit{Japanese-BCCWJ}. In total 1.51 million sentences are used in our experiments.

\paragraph{\indent Data Splits} The experiments in Section \ref{sec:experiments} use the 204k global test split. Initial comparisons were performed on the 915k dev set. The 102k training split was used to fine-tune \class{} and \boot{}. For early stopping, 31k sentences from the latter split were used as a held-out set. The exact instances are available in the associated code repository for future reproducibility.

\paragraph{\indent Genre Mapping} For 26 treebanks with instance-level genre labels in the metadata comments before each sentence, we created mappings from the treebank genre labels to the UD genre label set according to the guidelines described in Section \ref{sec:supervised-eval}. The genre metadata typically either follow the format \texttt{genre = X} or are implied by the document source specified in the sentence ID (e.g.\ \texttt{sent\_id = genre-...}). There are a total of 91 mappings which will be made available with the codebase upon publication.

\section{Model and Training Details}\label{sec:training-details}

The following describes architecture and training details for all methods. When not further defined, default hyperparameters are used. Implementations and predictions are available in the code repository at \href{https://personads.me/x/syntaxfest-2021-code}{\texttt{https://personads.me/x/syntaxfest-2021-code}}.

\paragraph{\indent Infrastructure} Neural models are trained on an NVIDIA A100 GPU with 40 GB of VRAM.

\paragraph{\indent Language Model} This work uses mBERT~\cite{devlin2019} as implemented in the Transformers library~\cite{wolf2020} as \texttt{bert-base-multilingual-cased}. Embeddings are of size $d_{\text{emb}}=768$ and the model has 178 million parameters. To create sentence embeddings, we use the mean-pooled WordPiece embeddings~\cite{wu2016} of the final layer.

\paragraph{\indent Classification} \class{} and \boot{} build on the standard mBERT architecture as follows: mBERT $\rightarrow$ \texttt{CLS}-token $\rightarrow$ linear layer ($d_{\text{emb}} \times 18$) $\rightarrow$ softmax. The training has an epoch limit of 30 with early stopping after 3 iterations without improvements on the development set. Backpropagation is performed using AdamW~\cite{loshchilov2017} with a learning rate of $10^{-7}$ on batches of size 16. The fine-tuning procedure requires GPU hardware which can host mBERT, corresponding to 10 GB of VRAM. Training on the 71k relevant instances takes approximately 10 hours.

\paragraph{\indent Clustering} Both \textit{Gaussian Mixture Models} (\textsc{GMM}) and \textit{Latent Dirichlet Allocation} (\citealp{blei2003}; \textsc{LDA}) use scikit-learn v0.23~\cite{sklearn}. \textsc{LDA} uses bags of character 3--6-grams which occur in at least 2 and in at most 30\% of sentences. GMMs use the mBERT sentence embeddings as input. Both methods are CPU-bound and cluster all treebanks in UD in under 45 minutes.

\paragraph{\indent Random Initializations} Each experiment is run thrice using the seeds 41, 42 and 43.

\section{Additional Results}\label{sec:additional-results}

Table \ref{tab:results-dev} shows results on the 915k development split of UD. Performance patterns are similar to those on the test split: the labeled clustering methods \gmml{} and \ldal{} perform best out of our proposed methods and outperform the baselines on the majority of metrics. With respect to classification, \boot{} outperforms both the noisier \class{} and \zero{}. Note that the frequency baseline \freq{} performs especially well on the dev set, since only 5 of 26 instance labeled treebanks are included and 4 of these have the majority genre \textit{news}.

\begin{table}[h!]
\centering
\resizebox{.48\textwidth}{!}{
\begin{tabular}{lrrrr}
\toprule
\textsc{Method} & \textsc{Pur} & \textsc{Agr} & \textsc{$\Delta$BC} & \textsc{F1} \\
\midrule
\freq & 100{\footnotesize$\pm$0.0} & 100{\footnotesize$\pm$0.0} & 23{\footnotesize$\pm$0.0} & 27{\footnotesize$\pm$0.0} \\
\zero & 43{\footnotesize$\pm$0.0} & 66{\footnotesize$\pm$0.0} & 50{\footnotesize$\pm$0.0} & 5{\footnotesize$\pm$0.0} \\
\midrule
\class & 87{\footnotesize$\pm$1.2} & 77{\footnotesize$\pm$3.9} & 29{\footnotesize$\pm$1.9} & 9{\footnotesize$\pm$4.5} \\ 
\boot & 95{\footnotesize$\pm$0.2} & 100{\footnotesize$\pm$0.0} & 24{\footnotesize$\pm$0.3} & 16{\footnotesize$\pm$1.0} \\ 
\midrule
\gmm & 92{\footnotesize$\pm$0.1} & 55{\footnotesize$\pm$5.5} & 30{\footnotesize$\pm$0.7} & ---\\
\footnotesize{\textsc{+Labels}} & 100{\footnotesize$\pm$0.0} & 100{\footnotesize$\pm$0.0} & 5{\footnotesize$\pm$0.1} & 17{\footnotesize$\pm$1.6} \\
\lda & 88{\footnotesize$\pm$1.0} & 42{\footnotesize$\pm$2.2} & 30{\footnotesize$\pm$0.2} & --- \\
\footnotesize{\textsc{+Labels}} & 100{\footnotesize$\pm$0.0} & 100{\footnotesize$\pm$0.0} & 5{\footnotesize$\pm$0.0} & 15{\footnotesize$\pm$0.9} \\
\bottomrule
\end{tabular}}
\caption{\label{tab:results-dev} \textbf{Results of Genre Prediction on UD (Dev).} Purity (\textsc{Pur} $\uparrow$), agreement (\textsc{Agr} $\uparrow$), overlap error ($\Delta$BC $\downarrow$) and micro-F1 over instance-labeled TBs (\textsc{F1} $\uparrow$) for \freq{}, \zero{}, \class{}, \boot{} and \gmm{}, \lda{} with/without labels. Standard deviation denoted $\pm$.}
\end{table}

\end{document}